%% file: main.tex
\documentclass[10pt,twocolumn,letterpaper]{article}

\usepackage{wacv}              
\usepackage{pifont}
\newcommand{\cmark}{\ding{51}} 
\newcommand{\xmark}{\ding{55}} 
\usepackage{adjustbox}
\usepackage{comment}
\usepackage{amsmath}

\definecolor{wacvblue}{rgb}{0.21,0.49,0.74}
\usepackage[pagebackref,breaklinks,colorlinks,allcolors=wacvblue]{hyperref}

\def\wacvPaperID{2477} 
\def\confName{WACV}
\def\confYear{2026}

\title{Disentangle and Regularize: Sign Language Production with Articulator-Based Disentanglement and Channel-Aware Regularization}

\author{
Sümeyye Meryem Taşyürek\thanks{meryemtasyurek@cs.hacettepe.edu.tr} \\
Hacettepe University
\and
Tuğçe Kızıltepe\thanks{tugcekiziltepe@hacettepe.edu.tr} \\
Hacettepe University
\and
Hacer Yalim Keles\thanks{hacerkeles@cs.hacettepe.edu.tr} \\
Hacettepe University
}

\begin{document}
\maketitle
\input{0_abstract}    
\input{1_intro}
\input{2_formatting}
\input{3_finalcopy}
{
    \small
    \bibliographystyle{ieeenat_fullname}
    \bibliography{main}
}

\end{document}

%% file: 0_abstract.tex
\begin{abstract}
In this work, we propose DARSLP, a simple gloss-free, transformer-based sign language production (SLP) framework that directly maps spoken-language text to sign pose sequences. We first train a pose autoencoder that encodes sign poses into a compact latent space using an articulator-based disentanglement strategy, where features corresponding to the face, right hand, left hand, and body are modeled separately to promote structured and interpretable representation learning. Next, a non-autoregressive transformer decoder is trained to predict these latent representations from word-level text embeddings of the input sentence. To guide this process, we apply channel-aware regularization by aligning predicted latent distributions with priors extracted from the ground-truth encodings using a KL divergence loss. The contribution of each channel to the loss is weighted according to its associated articulator region, enabling the model to account for the relative importance of different articulators during training. Our approach does not rely on gloss supervision or pretrained models, and achieves state-of-the-art results on the PHOENIX14T and CSL-Daily datasets.
\end{abstract}

%% file: 1_intro.tex
\section{Introduction}
\label{sec:intro}

Sign language is the primary mode of communication for Deaf and Hard-of-Hearing (DHH) individuals, involving a rich interplay of manual (hand movements) and non-manual (facial expressions, body posture) features. With the rise of deep learning, sign language production (SLP)—the task of generating sign sequences from spoken text—has gained increasing attention. However, producing natural and continuous sign articulation remains challenging due to its complex spatial-temporal dynamics.

Recent efforts focus on generating skeletal pose sequences as an intermediate representation, which offers a structured and efficient way to model articulation, separating motion synthesis from video rendering \cite{RASTGOO2024122846}. Yet, generating natural and expressive sign sequences remains an open challenge. Many models regress to mean poses, losing detail in hand and finger movements \cite{walsh2024signstitchingnovelapproach, ma-etal-2024-multi}, resulting in overly simplified gestures. Autoregressive models often suffer from error accumulation, leading to unnatural transitions \cite{DBLP:journals/corr/abs-2004-14874, DBLP:journals/corr/abs-2008-12405}. To address these issues, techniques such as adversarial training, progressive transformers, and mixture density networks (MDNs) have been explored \cite{DBLP:journals/corr/abs-2103-06982}, but still fall short in producing smooth and expressive hand articulations.

A significant research direction in SLP is reducing reliance on gloss annotations, intermediate labels between spoken and sign language. Although gloss-based models achieve strong performance, gloss annotation is costly, time-consuming, and often inconsistent across datasets, limiting scalability. To address this, gloss-free SLP has gained attention, targeting direct text-to-sign generation without gloss supervision \cite{10581980, walsh2024signstitchingnovelapproach, yin-etal-2024-t2s}. While promising, these models still face challenges in generating smooth transitions and expressive signing.

We propose DARSLP, a simple yet effective transformer-based SLP framework that directly translates spoken language text into continuous sign pose sequences without intermediate gloss annotations. The contribution of our work to the SLP domain can be summarized as follows:
\begin{itemize}
    \item We introduce an \textbf{articulator-based disentanglement} mechanism implemented via a pose autoencoder, where features corresponding to the face, right hand, left hand, and body are explicitly separated into distinct channel groups. This structured representation enables selective weighting of articulators (e.g., emphasizing hand movements) and facilitates the estimation of articulator-specific latent statistics. To the best of our knowledge, this straightforward yet effective inductive bias has not been previously explored in the context of SLP.

    \item We propose a \textbf{channel-aware regularization} strategy for training the non-autoregressive transformer by aligning predicted latent distributions with articulator-based channel priors derived from the autoencoder. In contrast to common variational methods that impose prior constraints, our approach directly computes these priors from observed channel distributions. Empirical results show that this regularization improves motion diversity and realism, and effectively reduces the regression-to-the-mean problem in generated pose sequences.
\end{itemize}

Unlike most prior methods that depend on gloss supervision, our model eliminates the need for gloss annotations while surpassing both gloss-based and gloss-free baselines in back-translation performance. It achieved third place in the CVPR 2025 SLRTP Challenge \cite{walsh2025slrtp}, demonstrating state-of-the-art performance in sign pose generation.

%% file: 2_formatting.tex
\section{Related Work}
\label{sec:related-work}

Sign language production (SLP) aims to generate sign sequences from spoken language text. Recently, studies have focused on producing skeletal pose sequences instead of full video synthesis, as this intermediate representation is more practical and generalizable \cite{RASTGOO2024122846}. Generating photorealistic videos from these poses remains a separate challenge. One of the primary challenges in continuous SLP, akin to continuous sign language translation (SLT), stems from the lack of one-to-one alignment between spoken language and sign language sequences. Realistic generation requires more than concatenating isolated signs; it demands smooth transitions and semantic coherence throughout the sequence. However, many SLP models still struggle to produce natural inter-sign dynamics.

\subsection{Autoregressive Approaches}

A major unsolved problem in SLP research is regression to the mean, a phenomenon in which generated sign sequences converge to average motion patterns, resulting in unnatural and overly simplified gestures \cite{walsh2024signstitchingnovelapproach}, \cite{ma-etal-2024-multi}. This leads to reduced expressiveness and semantic drift, lowering communication quality.

The field of SLP has seen significant advancements since the pioneering work of Stoll et al. (2018), who used neural machine translation (NMT) to map text to glosses, but failed to produce continuous, natural sign sequences \cite{StollStephanie20180903}. Later, Saunders \etal (2020a) introduced a transformer-based SLP model, but its autoregressive decoding amplified the regression-to-the-mean effect \cite{DBLP:journals/corr/abs-2004-14874}. To mitigate this, Saunders \etal (2020b) incorporated adversarial training with a multi-channel sign production framework. Although this approach systematically addressed the regression-to-the-mean problem for the first time, it did not fully resolve it \cite{DBLP:journals/corr/abs-2008-12405}. 

Later, methods such as progressive transformers and mixture density networks (MDN) have been explored, combined with data augmentation and adversarial training techniques \cite{DBLP:journals/corr/abs-2103-06982}. Despite these efforts, user evaluations indicate that the regression-to-the-mean problem persists. While MDN attempt to model the natural variation in sign sequences, it fails to capture fine-grained finger movements.

Our framework departs from autoregressive decoding entirely, eliminating sequential dependencies that amplify over-smoothing. By combining a non-autoregressive structure, which substantially reduces inference time  \cite{10.1145/3474085.3475463, hwang2022nonautoregressive, hwang2021non}, with a disentangled latent space, our approach addresses the regression-to-the-mean problem while producing expressive and natural motion without the need for adversarial objectives or complex multi-stage designs.

\label{sec:pose-autoencoder}
\begin{figure*}[htbp]
  \centering
  \includegraphics[width=\linewidth]{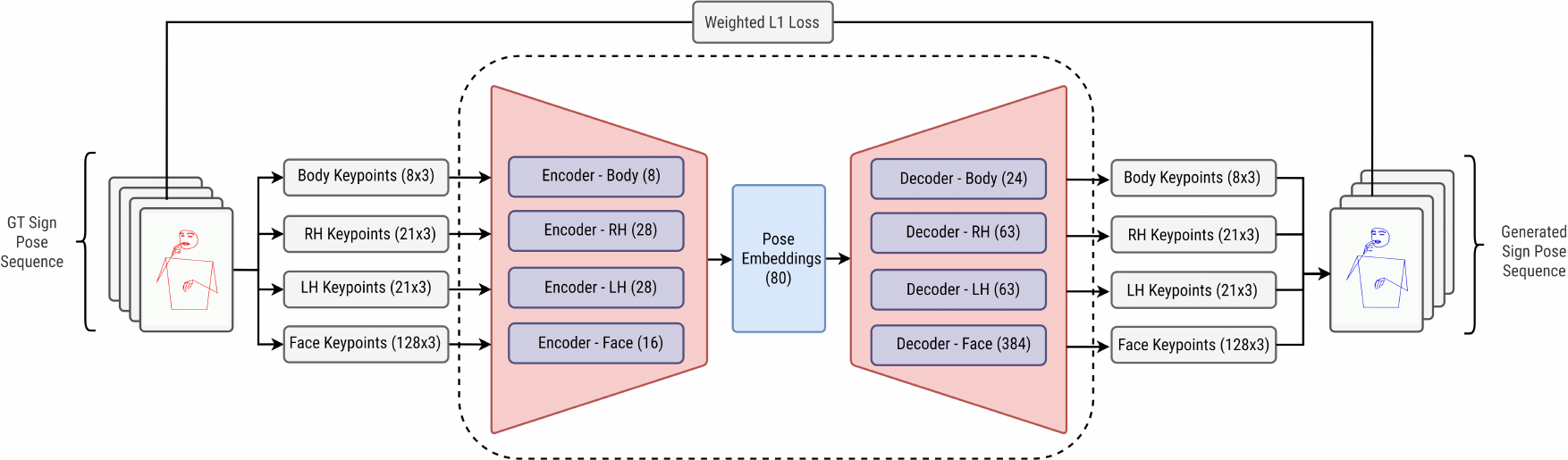}
  \caption{\textbf{Overview of the proposed pose autoencoder architecture.} Input 3D sign language poses are split into four regions: body (8 joints), right hand (21), left hand (21), and face (128). Each region is encoded separately to produce a compact 80-dimensional latent representation, then decoded by region-specific decoders to reconstruct the full pose sequence. A weighted L1 loss emphasizes critical articulators, such as the hands, during training. 
  }
  \label{fig:ae-architecture}
\end{figure*}

\subsection{Non-Autoregressive Approaches}

To address limitations in autoregressive decoding, Hwang \etal (2021) proposed NSLP-G, which used a Variational Autoencoder (VAE) to generate sign poses in a Gaussian latent space \cite{hwang2021non}. However, it struggled with sequence length prediction and failed to capture fine-grained finger movements. A follow-up model introduced length regulators and replaced MSE with BCE loss, improving low-variance detail modeling \cite{hwang2022nonautoregressive}. Still, hand articulation remained poorly encoded, as VAEs tended to emphasize larger body motions and overlook fine manual features.

Ma \etal (2024) tackled the regression-to-the-mean problem by introducing a dual-decoder transformer, separating manual and full-body decoding \cite{ma-etal-2024-multi}. Although this improved hand modeling, the approach suffered from high computational cost and sensitivity to data quality, limiting its scalability and generalization.

Our work advances non-autoregressive SLP by introducing a semantically structured latent space that explicitly disentangles hands, face, and body. This factorization allows for finer modeling of manual and non-manual features while maintaining computational efficiency. Furthermore, our length prediction mechanism provides robust temporal control without instability.

\subsection{Gloss-Free Approaches}

While gloss-based SLP models yield high accuracy, gloss annotation is costly and time-consuming, limiting scalability. To address this, gloss-free approaches aim to translate spoken language directly into sign language sequences without gloss labels. Hwang \etal (2024) proposed SignVQNet, converting pose sequences into discrete tokens within an autoregressive structure \cite{10581980}. Walsh \etal (2024) also leveraged vector quantization (VQ) to reduce gloss dependency \cite{walsh2024signstitchingnovelapproach, walsh2024datadrivenrepresentationsignlanguage}. However, VQ-based models often produce unnatural transitions and lose fine-grained motion detail. Yin \etal (2024) addressed this with Dynamic VQ (DVQ-VAE), which adapts encoding length to information density \cite{yin-etal-2024-t2s}, but challenges with smooth transitions and autoregressive error accumulation persist.

Alternative symbolic systems like HamNoSys \cite{Hanke2004HamNoSysR} and SignWriting \cite{sutton2022lessons} aim to replace glosses by encoding articulatory or visual features. These have shown promise in tasks like Text-to-HamNoSys (T2H) \cite{walsh2022changingrepresentationexamininglanguage} and SignWriting-based synthesis \cite{inproceedingsBouzid, jiang2023machinetranslationspokenlanguages}, but still require manual annotation and lack scalability.

In contrast, our work adopts a fully gloss-free approach that learns to generate continuous sign poses directly from spoken text, enabling end-to-end, scalable SLP without intermediate linguistic supervision.

\subsection{Latent Space Modeling and Disentanglement}

Beyond discrete tokenization, several studies have explored continuous latent space modeling using autoencoders or VAEs \cite{hwang2021non, hwang2022nonautoregressive}. These methods aim to learn compact embeddings capturing both manual and non-manual features. However, they typically treat pose data holistically, leading to inefficient encoding of fine-grained articulations, particularly for hands and fingers, while favoring global motion patterns.

Outside the SLP domain, autoencoders have demonstrated success in learning structured representations across modalities, supporting tasks such as sequence modeling \cite{dai2015semisupervisedsequencelearning}, biomedical analysis \cite{Way174474}, and latent variable modeling \cite{kingma2022autoencodingvariationalbayes}. In multimodal settings, disentangling representations is key to modeling distinct signal sources while preserving coherence.

Motivated by this, we design a structurally disentangled autoencoder tailored to the multimodal nature of sign language, separating latent representations for hands, face, and body. Unlike prior SLP works that rely on holistic or discrete representations, our method explicitly captures the dynamics of each articulator group.

Additionally, we leverage KL divergence as a targeted regularization signal over these disentangled channels. While widely used in VAEs for latent control \cite{DBLP:journals/corr/BowmanVVDJB15}, KL loss remains underexplored in SLP. We apply it outside a variational setting to enforce structural coherence across regions, improving representation smoothness without sacrificing detail.

\section{Methodology}
\label{sec:method}

\begin{figure*}[htbp]
  \centering
  \includegraphics[width=\linewidth]{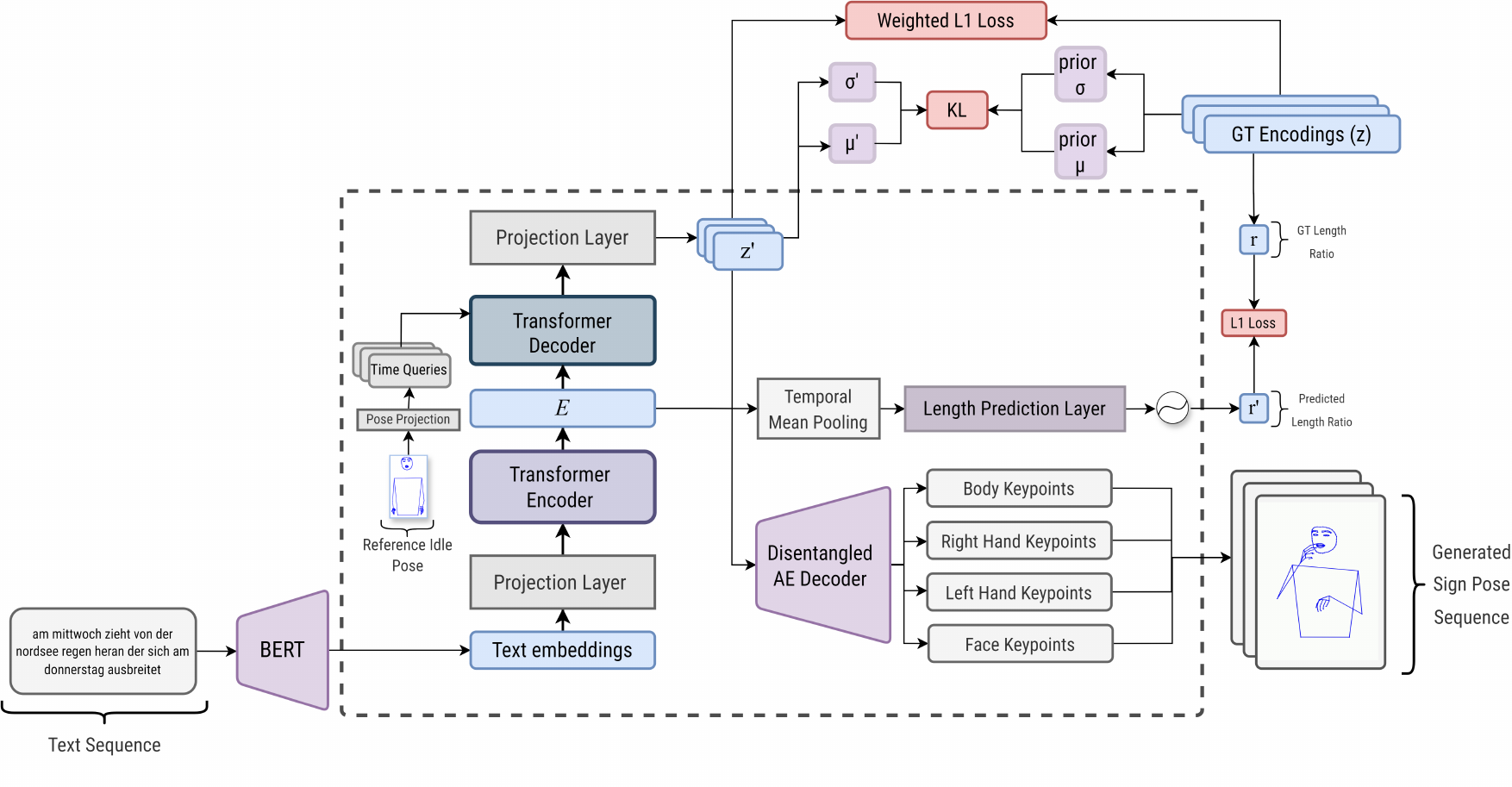}
  \caption{\textbf{Overview of the proposed DARSLP framework.} Input text is encoded with a pre-trained BERT model to obtain sentence-level embeddings, which are processed by non-autoregressive transformer to predict pose latents in the disentangled latent space. A temporal mean-pooling supports length prediction, while the decoder is conditioned on time queries derived from an idle pose. Training uses a region-specific weighted L1 loss and KL term for regularization. The disentangled AE decoder then reconstructs regions keypoints to produce the final sign pose sequence.}
  \label{fig:nar-architecture}
\end{figure*}

\subsection{Pose Autoencoder}
To model the multimodal structure of sign language articulation, we design a structurally disentangled pose autoencoder that encodes and reconstructs sign poses within a compact latent space. Rather than aiming to achieve disentanglement in the strict unsupervised sense, we explicitly incorporated an architectural design that encourages the model to learn semantically organized representations. This design introduces a multi-modal latent factorization that aligns well with the physical structure and functional roles of each body part in sign language expression. 

Our model processes 3D skeleton sequences comprising 8 upper body and 21 left hand, 21 right hand joints, along with 128 facial keypoints, each represented by (x, y, z) coordinates. As shown in Figure \ref{fig:ae-architecture}, the encoder is anatomically partitioned into four regions, upper body, right hand, left hand, and face, each handled by a dedicated linear encoder. The pose autoencoder takes these 3D joint coordinates arranged in a fixed, anatomically consistent order defined by the dataset’s skeleton topology (e.g., Mediapipe format). 

Based on ablation results (Section \ref{sec:ablation-latent}), we fix the total latent space to 80 dimensions, assigning 16 to the face and distributing the remaining 64 across other regions in proportion to their joint counts; Upper body (8 joints): mapped from 8 × 3 to 8 dimensions, Right/Left hand (21 joints each): mapped from 21 × 3 to 28 dimensions per hand, Face (128 joints): mapped from 128 × 3 to 16 dimensions. Although the face has high raw input dimensionality, many of its keypoints—particularly in the cheeks, jawline, and forehead—exhibit correlated motion. This redundancy allows the encoder to use fewer latent dimensions for the face without significantly affecting reconstruction quality.

To accommodate the characteristics of each dataset, we use slightly different autoencoder architectures. For PHOENIX14T \cite{8578910}, each articulator is encoded using a single linear projection layer, allowing direct mapping from raw 3D joint inputs to their respective latent representations. This simple and efficient structure is sufficient for PHOENIX14T, which has a more constrained linguistic domain and relatively lower articulatory variability. The decoder modules also consist of single linear layers per region, mirroring the encoder structure. For CSL-Daily \cite{zhou2021signbt}, which involves broader linguistic context and higher articulatory variability, we introduce an additional projection layer with nonlinearity for Right/Left hand and face regions. Specifically, each region  is encoded using a two-layer MLP comprising a linear projection, a PReLU activation, and a second linear layer. This increased representational capacity helps stabilize the latent space, particularly for hand and face regions where variability is highest. The intermediate hidden dimension is set to 40 for both hands and 96 for the face. Decoder modules follow the same structure, using symmetric two-layer MLPs with matching intermediate widths.

\subsection{Transformer Model}
\label{sec:transformer-model}

Following the latent pose representation learned by the pose autoencoder, the second part of our framework generates sign pose embeddings conditioned on spoken language text. We utilize a Transformer-based architecture that maps sentence-level text to the latent space learned by the pose autoencoder.

Our model, as illustrated in Figure \ref{fig:nar-architecture}, uses a Transformer encoder-decoder architecture. Input text to the encoder is represented by 768-dimensional BERT-based word embeddings, which are linearly projected to 512-dimension to match the model’s internal dimensions. The encoder consists of 3 layers with 4 attention heads and 1024-dimensional feed-forward networks, along with positional encoding to retain temporal ordering of the text sequence. 

The decoder follows a non-autoregressive structure to reduce error accumulation and predicts 80-dimensional latent vectors corresponding to the autoencoder’s space. It includes 6 layers with 8 attention heads and the same 1024-dimensional feed-forward size. Temporal dynamics are initialized using learned time queries derived from a fixed stationary reference pose, where both hands rest downward as when signers are not performing a sign. This pose is projected into the decoder space and expanded across all time steps to provide a consistent initialization for sequence generation.

Ground truth 80-dimensional pose encodings are extracted using the pretrained autoencoder and serve as structured supervision for transformer training. These encodings provide structurally disentangled representations of manual (hands) and non-manual (face, upper body) articulators, guiding the model in learning mappings from text to pose. To ensure alignment with the autoencoder’s latent space, we incorporate L1 and KL-based regularization during training, encouraging accurate and coherent pose generation.

We also incorporate a length predictor to dynamically adjust the length of the generated sign sequences based on the given input sentences. First, a temporal mean pooling is applied, averaging across the temporal dimension while ignoring padding positions. This sentence-level embedding from the encoder is passed through a linear layer with sigmoid activation to predict a normalized length ratio (0–1), which corresponds to the proportion of the maximum possible sequence length required to represent the given input sentence. The module is trained using supervision from ground-truth sequence lengths, promoting temporal alignment consistent with semantic load, in line with the prior work \cite{yin-etal-2024-t2s}. 

The predicted latent sequence is decoded into 3D pose keypoints using the pretrained pose autoencoder's decoder. This design enables efficient, end-to-end sign pose generation without the need for additional refinement or post-processing.

\subsection{Training Procedure}
\label{sec:training-procedure}

Our training follows a two-stage strategy for mapping spoken language to continuous sign pose sequences. In the first stage, we train a structurally disentangled autoencoder to learn compact latent representations that separately encode manual (hands) and non-manual (face, upper body) features. In the second stage, a non-autoregressive Transformer is trained to predict these latent representations from word-level text embeddings of the input sentence, supervised by the ground-truth encodings from the pretrained autoencoder.

\subsubsection{Training the Pose Autoencoder}
\label{sec:training-ae}

The structurally disentangled autoencoder is trained to minimize a \textit{weighted reconstruction loss} across four anatomical regions, body (B), right hand (RH), left hand (LH), and face (F) combined with \textit{L1 regularization} applied only to encoder weights to promote sparsity:

{\footnotesize
\begin{equation}
\mathcal{L}_{\text{ae}} = \sum_{R \in \{\text{B},\,\text{RH},\,\text{LH},\,\text{F}\}} 
w_R \frac{1}{N_R} \sum_{i=1}^{N_R} \|\hat{\mathbf{x}}_R^{(i)} - \mathbf{x}_R^{(i)}\|_{1}
+ \lambda \sum_{j \in \{\text{enc}\}} \| \mathbf{W}_j \|_{1}
\label{eq:enc-total}
\end{equation}
}

\noindent
Here, $\mathbf{x}_R^{(i)}$ and $\hat{\mathbf{x}}_R^{(i)}$ denote the ground truth and reconstructed keypoints for region $R$ at frame $i$, and $N_R$ is the number of frames. The first term is a weighted L1 reconstruction loss, where the weights $w_R$ allow articulator-specific emphasis during training. The second term is an L1 regularization loss applied solely to the encoder weights, a strategy used in sparse autoencoders \cite{10.5555/1756006.1953039}. This preserves the decoder’s full capacity to reconstruct fine-grained spatial dynamics, which is particularly critical for manual articulators like fingers.

\subsubsection{Training Generator Model}
\label{sec:training-generator}

The training objective of our transformer model is divided into two phase to effectively balance the learning of accurate pose encodings and structured latent space distribution. In the first-phase, the model is trained to predict ground truth (GT) pose encodings obtained from the pretrained pose autoencoder. In this phase, we minimize an L1 loss between the predicted encodings for the transformer model and the obtained GT encodings.

{\footnotesize
\begin{equation}
\mathcal{L}_{\text{phase1}} = \sum_{R \in \{\text{B}, \text{RH}, \text{LH}, \text{F}\}} w_R \cdot \frac{1}{T} \sum_{t=1}^{T} \left\| \hat{z}_t^{(R)} - z_t^{(R)} \right\|_1 
+ \left\| \hat{r} - r \right\|_1
\label{eq:phase1-loss}
\end{equation}
}

\noindent
Here, \( \hat{z}_t^{(R)} \) and \( z_t^{(R)} \) denote the predicted and ground-truth latent embeddings for region \( R \) at timestep \( t \), and \( w_R \) are region-specific weights. The second term supervises the predicted sequence length ratio \( \hat{r} \) using the ground-truth \( r \). We will refer to this model as \textbf{DARSLP}.

In the second phase of the training, we incorporate a KL divergence term to promote generalization and preserve the distributions of the structurally disentangled latent space. We refer to the resulting model as \textbf{DARSLP-KL}, which extends the training of the first-phase model by adding regularization on the channel distributions (Eqn. \ref{eq:loss-phase2}). In this context, during training, we enforce the predicted latent channel distributions to align with precomputed priors derived from ground truth encodings for each articulator region. 

{\scriptsize
\begin{equation}
\mathcal{L}_{\text{phase2}} = \mathcal{L}_{\text{phase1}} + \sum_{R \in \{\text{B}, \text{RH}, \text{LH}, \text{F}\}} \sum_{c \in R} D_{\text{KL}}\left( \mathcal{N}(\mu^{(c)}, \sigma^{(c)}) \,\|\, \mathcal{N}(\mu_{\text{prior}}^{(c)}, \sigma_{\text{prior}}^{(c)}) \right)
\label{eq:loss-phase2}
\end{equation}
}

\noindent
Here, \( \mu^{(c)} \) and \( \sigma^{(c)} \) denote the batch-wise mean and variance of predicted latent channels, while \( \mu_{\text{prior}}^{(c)} \) and \( \sigma_{\text{prior}}^{(c)} \) are prior statistics computed from the autoencoder’s training embeddings. 


%% file: 3_finalcopy.tex
\section{Experiments}
\subsection{Implementation Details}
\label{sec:experiments}

\noindent\textbf{Datasets.} We evaluate our model on two continuous sign language datasets. PHOENIX-2014T \cite{8578910} contains 8,247 German Sign Language sentences aligned with gloss and spoken German. We use the 3D pose annotations from the CVPR 2025 SLRTP Challenge \cite{walsh2025slrtp}, where 2D keypoints extracted via MediaPipe Holistic \cite{lugaresi2019mediapipeframeworkbuildingperception} are uplifted to 3D using Ivashechkin \etal’s method \cite{ivashechkin2023improving3dposeestimation}, yielding $N \times 178 \times 3$ tensors. CSL-Daily \cite{zhou2021signbt} includes 20,654 Chinese Sign Language (CSL) sentences from 10 signers. We extract 3D poses using MediaPipe Holistic, standardizing each frame to 178 keypoints (upper body, hands, face). To ensure consistency across signers and datasets, poses are normalized by centering on the neck and scaling by shoulder width. \newline

\noindent\textbf{Autoencoder training settings.} We assign region-specific weights to the reconstruction loss: 
$w_{\text{RH}} = w_{\text{LH}} = 1.5$,
$w_{\text{F}} = 1.0$,
$w_{\text{B}} = 0.5$.
These weights prevent high-variance upper body motions from overshadowing fine-grained hand and facial cues. L1 regularization 
with $\lambda = 1 \times 10^{-4}$ promotes encoder sparsity. Optimization is performed using the Adam optimizer, with a learning rate of \( 2 \times 10^{-4} \) and beta parameters set to \( (0.5, 0.9) \). The pose autoencoder is trained for 270 epochs on both PHOENIX14T and CSL-Daily datasets. \newline

\noindent\textbf{Transformer training settings.} For L1 loss, we use;
$w_{\text{RH}} = 14$, $w_{\text{LH}} = 10$, $w_{\text{F}} = 2$, decided based on ablation studies (Section \ref{sec:ablation-transformer}) comparing different weighting schemes. Hands carry the core lexical content in sign language, with the right hand as dominant in DGS and CSL, conveying most meaning, and the left hand serving a supportive role. The body offers coarse spatial context, while the face encodes grammatical and emotional cues but involves redundant motion across many keypoints. We apply similar region-based weighting in both the autoencoder and transformer training settings to reflect these functional roles. Regions are weighted accordingly, guided by ablation results.

We employ the Adam optimizer with a learning rate of \(2 \times 10^{-4}\), weight decay of \(1 \times 10^{-4}\), and a ReduceLROnPlateau scheduler (factor 0.9, patience 40) in both phases of the training. Early stopping based on validation loss is applied to prevent overfitting. Training is performed on a 4×NVIDIA A100-SXM4-40GB setup using PyTorch Lightning \cite{Falcon_PyTorch_Lightning_2019}.
\subsection{Ablation Study on Latent Dimensionality}
\label{sec:ablation-latent}

We first analyze how different allocations of latent dimensions across body, hands, and face affect performance. Table~\ref{tab:latent-dim-ablation} compares multiple configurations, keeping the body+hands dimension fixed or varied and adjusting the face block. Results are reported on the PHOENIX14T development set for all ablations.

\begin{table}[h]
\centering
\scriptsize
\setlength{\tabcolsep}{4pt}
\begin{tabular}{lcccccc}
\toprule
\textbf{Latent Dim Config.} & \textbf{B1↑} & \textbf{B2↑} & \textbf{B3↑} & \textbf{B4↑} & \textbf{chrF↑} & \textbf{ROUGE↑} \\
\midrule
32 hands\&body + 8 face               & 20.79 & 11.81 & 8.34 & 6.43 & 27.01 & 18.44 \\
64 hands\&body + 8 face               & 22.54 & 13.84 & 9.75 & 7.44 & 28.58 & 21.18 \\
128 hands\&body + 8 face              & 20.44 & 11.99 & 8.45 & 6.54 & 27.43 & 18.67 \\
64 hands\&body + 16 face              & \textbf{24.07} & \textbf{14.25} & 9.73 & 7.25 & 28.51 & 21.30 \\
64 hands\&body + 32 face              & 23.04 & 14.00 & \textbf{9.88} & \textbf{7.54} & \textbf{28.67} & \textbf{21.42} \\
64 hands\&body + 64 face              & 20.75 & 12.39 & 8.72 & 6.67 & 27.71 & 19.07 \\
\textit{80-dim entangled (no split)} & 19.61 & 11.34 & 7.97 & 6.13 & 27.21 & 18.10 \\
\bottomrule
\end{tabular}
\caption{Effect of different latent dimension configurations.}
\label{tab:latent-dim-ablation}
\end{table}


The ablation results in Table~\ref{tab:latent-dim-ablation} show that the best overall performance is achieved with 64 latent dimensions allocated to body and hands, distributed proportionally to joint counts (8 for body, 28 for each hand) to preserve anatomical consistency. For the face, despite having a much larger number of keypoints (128), we observe that 8, 16, and 32 latent dimensions all produce very similar results. This indicates strong interdependence among facial keypoints, allowing the face representation to be effectively compressed into a compact latent block without loss of information. Furthermore, all disentangled models outperform the entangled configuration with an 80-dimensional flat latent vector in BLEU-4 and ROUGE metrics, highlighting the benefit of factorized and structured representations.

\subsection{Ablation Study on Generator Training}
\label{sec:ablation-transformer}

We next analyze the impact of region-specific L1 loss weighting on generator training. In this context, the three numbers in ($w_{R\text{H}} - w_{L\text{H}} - w_F$) indicate the weighting coefficients assigned to the right hand, left hand, and face regions, respectively. As shown in Table~\ref{tab:l1_weight_ablation}, moderate weighting of the hands, such as (5-5-1), consistently improves performance over the unweighted baseline (1-1-1), demonstrating the benefit of emphasizing manual components. Further gains are achieved by prioritizing the dominant right hand, with the best results observed at the (14-10-2) configuration for the 64H\&B + 16F setup. Increasing the weights beyond this point causes a clear performance drop.

\begin{table}[h]
\centering
\scriptsize
\setlength{\tabcolsep}{7pt}
\begin{tabular}{lcccccc}
\toprule
\textbf{RH-LH-F} & \textbf{B1↑} & \textbf{B2↑} & \textbf{B3↑} & \textbf{B4↑} & \textbf{chrF↑} & \textbf{ROUGE↑} \\
\midrule
\multicolumn{7}{c}{\textbf{64H\&B + 32F}} \\
\midrule
(1-1-1)        & 23.04 & 14.00 & 9.88 & 7.54 & 28.67 & 21.42 \\
\textbf{(5-5-1)}   & \textbf{23.64} & \textbf{14.90} & \textbf{10.50} & \textbf{8.13} & \textbf{29.52} & \textbf{22.30} \\
(7-5-1)       & 20.76 & 12.25 & 8.44 & 6.46 & 27.51 & 19.36 \\
(14-10-2)     & 20.55 & 12.58 & 8.86 & 6.79 & 27.19 & 18.93 \\
(21-15-2)     & 19.92 & 12.20 & 8.50 & 6.40 & 26.72 & 18.12 \\
\midrule
\multicolumn{7}{c}{\textbf{64H\&B + 16F}} \\
\midrule
(1-1-1)       & 24.07 & 14.25 & 9.73 & 7.25 & 28.51 & 21.30 \\
(7-5-1)       & 23.96 & 15.16 & 10.68 & 8.06 & 29.23 & 22.48 \\
(14-10-1)     & 24.18 & 15.47 & 11.00 & 8.50 & 29.52 & 22.81 \\
\textbf{(14-10-2)} & \textbf{24.97} & \textbf{15.91} & \textbf{11.35} & \textbf{8.72} & \textbf{29.67} & \textbf{23.25} \\
(14-10-4)     & 22.85 & 13.82 & 9.49 & 7.14 & 27.91 & 21.21 \\
(21-15-2)     & 24.51 & 15.18 & 10.59 & 7.88 & 29.27 & 22.30 \\
\midrule
\multicolumn{7}{c}{\textbf{64H\&B + 8F}} \\
\midrule
(1-1-1)       & 22.54 & 13.84 & 9.75 & 7.44 & 28.58 & 21.18 \\
\textbf{(5-5-1)} & \textbf{23.03} & \textbf{14.46} & \textbf{10.38} & \textbf{8.03} & \textbf{29.08} & \textbf{21.62} \\
(7-5-1)       & 22.18 & 13.73 & 9.53 & 7.19 & 27.90 & 20.78 \\
(14-10-2)     & 21.59 & 13.47 & 9.44 & 7.16 & 27.47 & 19.27 \\
(21-15-2)     & 20.98 & 12.88 & 8.90 & 6.80 & 26.84 & 18.70 \\
\bottomrule
\end{tabular}
\caption{Effect of varying L1 loss weights for right hand (RH), left hand (LH), and face (F) components while keeping body+hands fixed at 64 dimensions. Bold entries indicate optimal configurations before degradation sets in.}
\label{tab:l1_weight_ablation}
\end{table}

We further investigate the impact of KL regularization and its scheduling. As shown in Table~\ref{tab:KL-results}, omitting KL results in suboptimal performance, while applying it from the start causes instability. Our two-phase strategy, where the model is first trained with only L1 loss and then fine-tuned with KL, achieves the best results across all metrics. This supports prior findings in VAE literature \cite{DBLP:journals/corr/BowmanVVDJB15, sønderby2016laddervariationalautoencoders, fu-etal-2019-cyclical} that KL regularization must be carefully scheduled. While our model does not follow a variational formulation, the findings emphasize that applying regularization in a staged manner can still play a critical role in enforcing structure within the latent space.


\begin{table}[h]
\centering
\scriptsize
\setlength{\tabcolsep}{5pt}
\begin{tabular}{lccccccc}
\toprule
\textbf{Model} & \textbf{B1↑} & \textbf{B2↑} & \textbf{B3↑} & \textbf{B4↑} & \textbf{chrF↑} & \textbf{ROUGE↑} \\
\midrule
GT (dev) & 30.92 & 21.34 & 16.01 & 12.74 & 34.71 & 30.19 \\
\midrule
L1 w/o KL & 24.97 & 15.91 & 11.35 & 8.72 & 29.67 & 23.25 \\
L1 + w/ KL & 20.78 & 12.64 & 9.01 & 6.98 & 27.92 & 19.66 \\
\textbf{Two-Phase (Ours)} & \textbf{28.41} & \textbf{19.31} & \textbf{14.45} & \textbf{11.40} & \textbf{31.85} & \textbf{26.93} \\							
\bottomrule
\end{tabular}
\caption{Effect of KL divergence on training strategies using 80-dim latent space and (14-10-2) L1 weighting.}
\label{tab:KL-results}
\end{table}

\subsection{Latent Space Analysis}

To isolate and interpret the contribution of each articulator within the latent spaces of both entangled and disentangled autoencoders, we adopt a targeted visualization strategy on the PHOENIX14T dataset. For each input sample, only the keypoints of a selected region (e.g., right hand) are retained, while all other keypoints are replaced with a canonical reference pose that is computed as the mean pose over the training set. This masking ensures that variation in the latent space is driven solely by the selected region, while the remaining parts of the body remain fixed, minimizing confounding effects and providing a stable reference frame. The masked sequences are then passed through the model to obtain 80-dimensional latent representations, which are standardized using training statistics for visualization. 

Principal Component Analysis (PCA) \cite{WOLD198737} is first applied to the latent representations to suppress noise and retain the most informative variance components. This is followed by Uniform Manifold Approximation and Projection (UMAP) \cite{mcinnes2020umapuniformmanifoldapproximation}, which maps the PCA-reduced features to a 2D space for clearer visualization. Both PCA and UMAP are fitted on the training set and consistently applied to the development set to ensure comparability. This unified analysis pipeline allows fair comparison between entangled and disentangled autoencoders. In particular, masking is crucial for the entangled AE, as it does not maintain region-specific latent channels and would otherwise produce entangled activations influenced by irrelevant articulators.

\begin{figure}[htbp]
    \centering
    \begin{minipage}[b]{0.49\columnwidth}
        \centering
        \includegraphics[width=\textwidth]{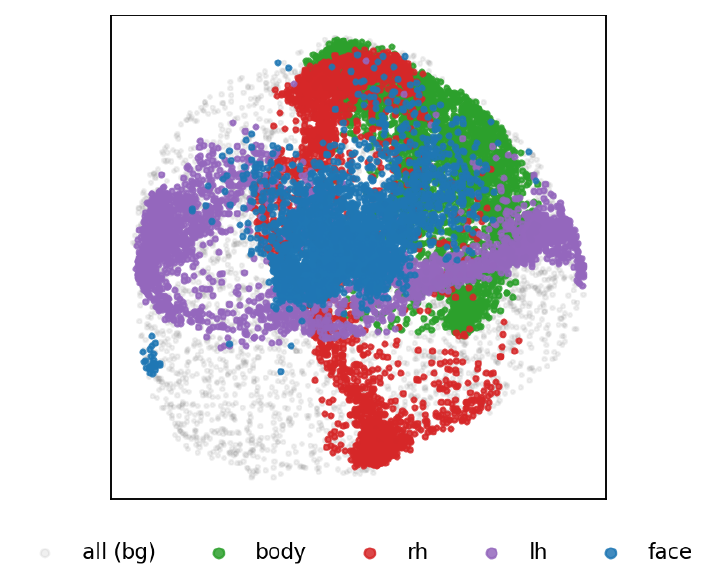}
        \vspace{0.5mm}
        \small\textbf{(a)} Entangled AE
    \end{minipage}
    \hfill
    \begin{minipage}[b]{0.49\columnwidth}
        \centering
        \includegraphics[width=\textwidth]{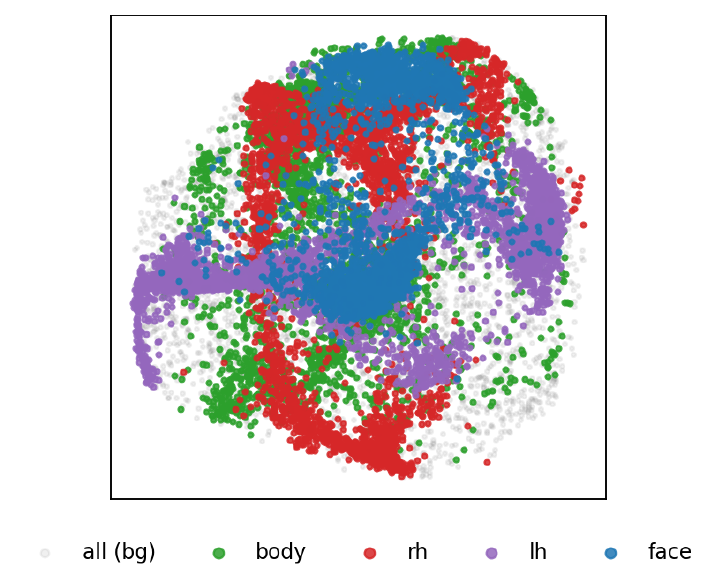}
        \vspace{0.5mm}
        \small\textbf{(b)} Disentangled AE
    \end{minipage}

    \caption{\textbf{Masked UMAP comparison of entangled vs. disentangled AE.} Colors represent different regions; gray shows unmasked latents.}
    \label{fig:entangled_vs_disentangled_ae}
\end{figure}

In the entangled AE visualization, the body and face are represented as compact, near-unimodal clusters with substantial overlap between different articulators (Fig. \ref{fig:entangled_vs_disentangled_ae}(a)). In contrast, the disentangled AE preserves the intrinsic variance of each region, resulting in more structured and well-separated latent distributions \ref{fig:entangled_vs_disentangled_ae}(b)). The hands and face, in particular, display rich and complex multi-modal patterns that reflect the fine-grained dynamics of sign language movements. This separation reveals underlying structures in the data that are suppressed in the entangled representation, indicating that the disentangled AE is more effective at capturing region-specific motion patterns critical for sign language production.

In addition to analyzing the latent space of autoencoders, we further examine how the latent representations produced by our Transformer-based generator compare to those of the AE and how KL regularization influences these representations. To this end, we perform a separate PCA analysis for each articulator. Latent channels corresponding to the body, right hand, left hand, and face are standardized using region-specific training statistics, and distinct PCA models are fitted on each region’s training data to obtain independent 2D projections.

\begin{figure}[htbp]
    \centering

    \begin{minipage}[b]{0.30\columnwidth}
        \centering
        \includegraphics[width=\textwidth]{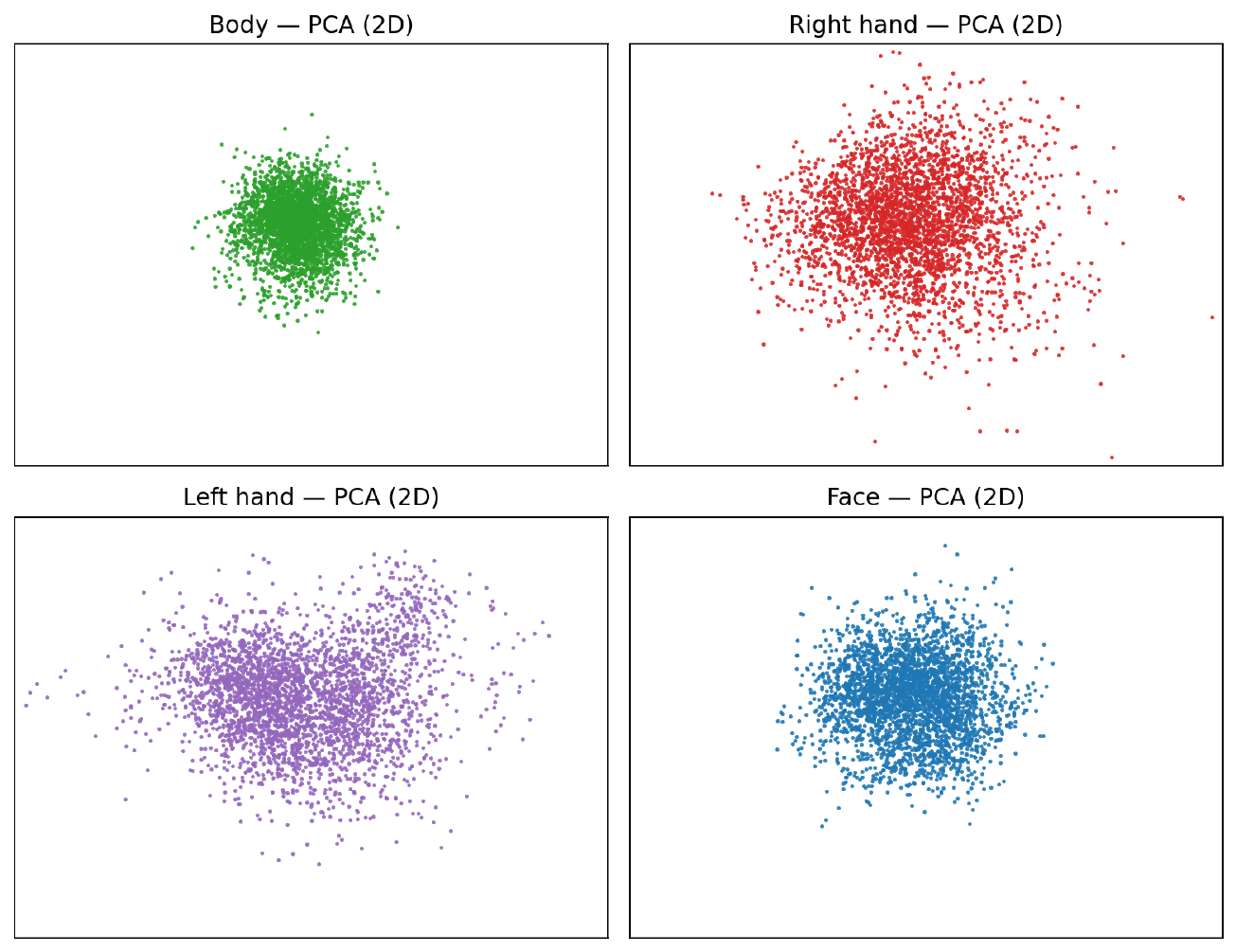}
        \vspace{1mm}
        \small\textbf{(a)} GT regional PCA
    \end{minipage}
    \hfill
    \begin{minipage}[b]{0.30\columnwidth}
        \centering
        \includegraphics[width=\textwidth]{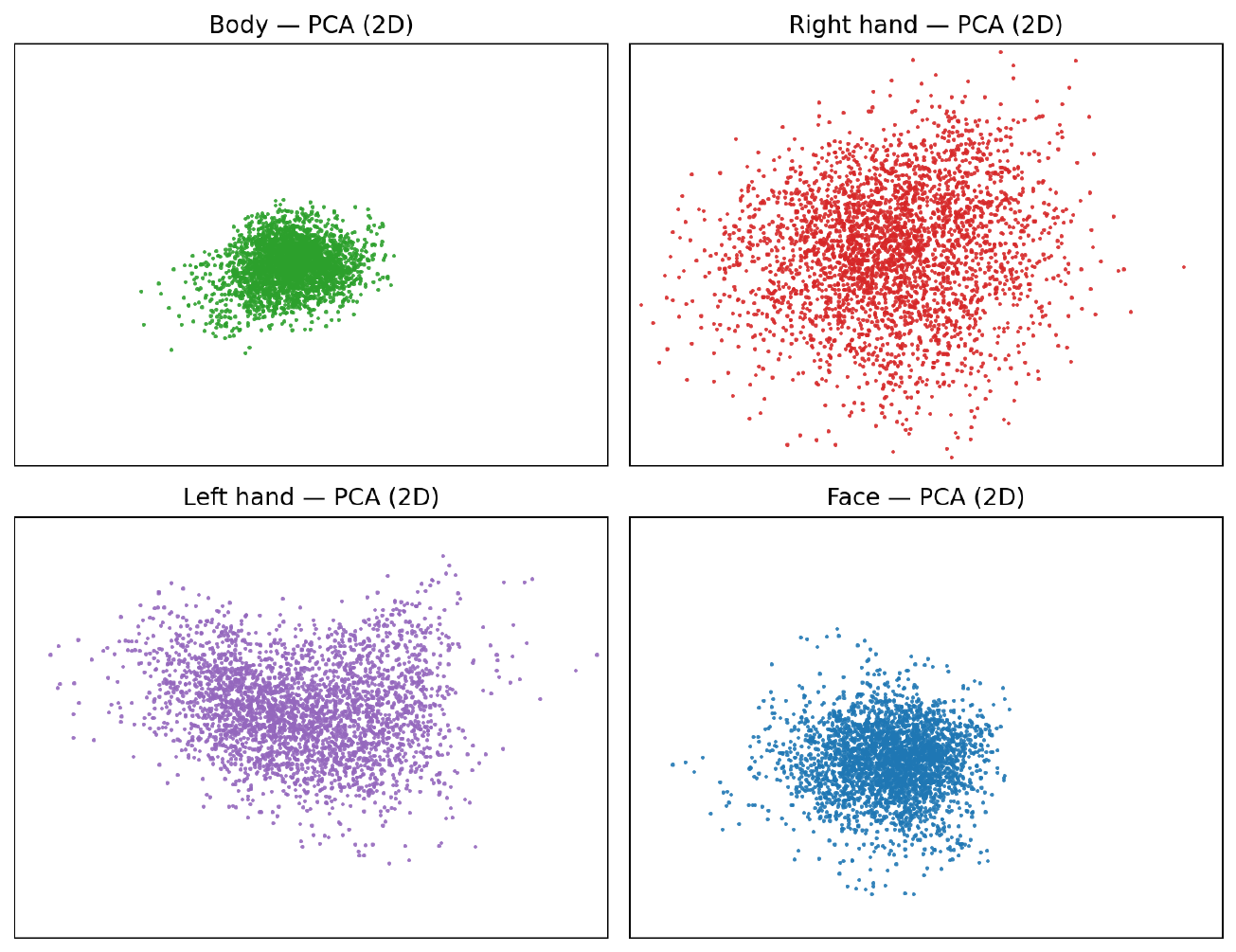}
        \vspace{1mm}
        \small\textbf{(b)} Transformer w/o KL
    \end{minipage}
    \hfill
    \begin{minipage}[b]{0.30\columnwidth}
        \centering
        \includegraphics[width=\textwidth]{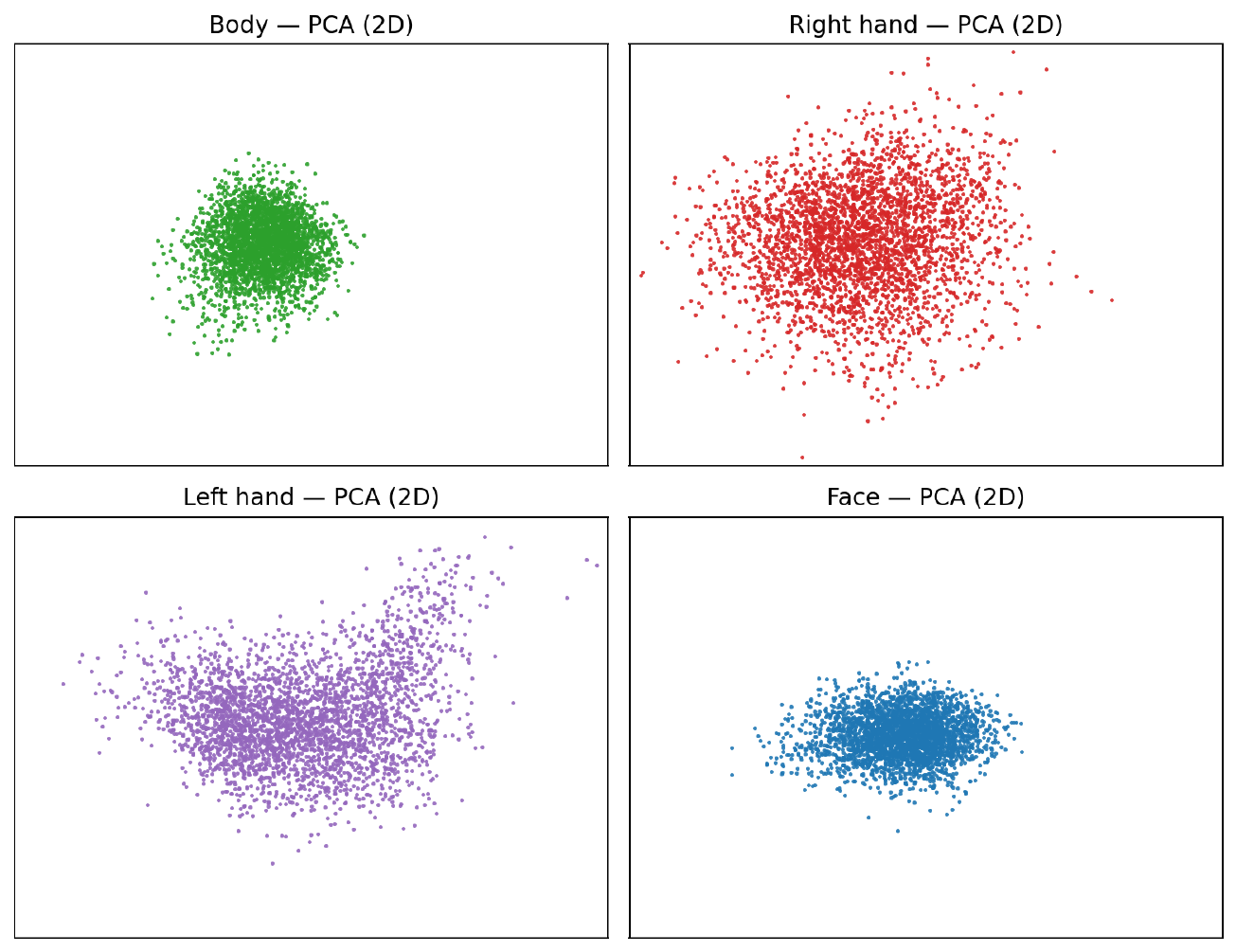}
        \vspace{1mm}
        \small\textbf{(c)} Transformer w/ KL
    \end{minipage}

    \vskip\baselineskip

    \begin{minipage}[b]{0.46\columnwidth}
        \centering
        \includegraphics[width=\textwidth]{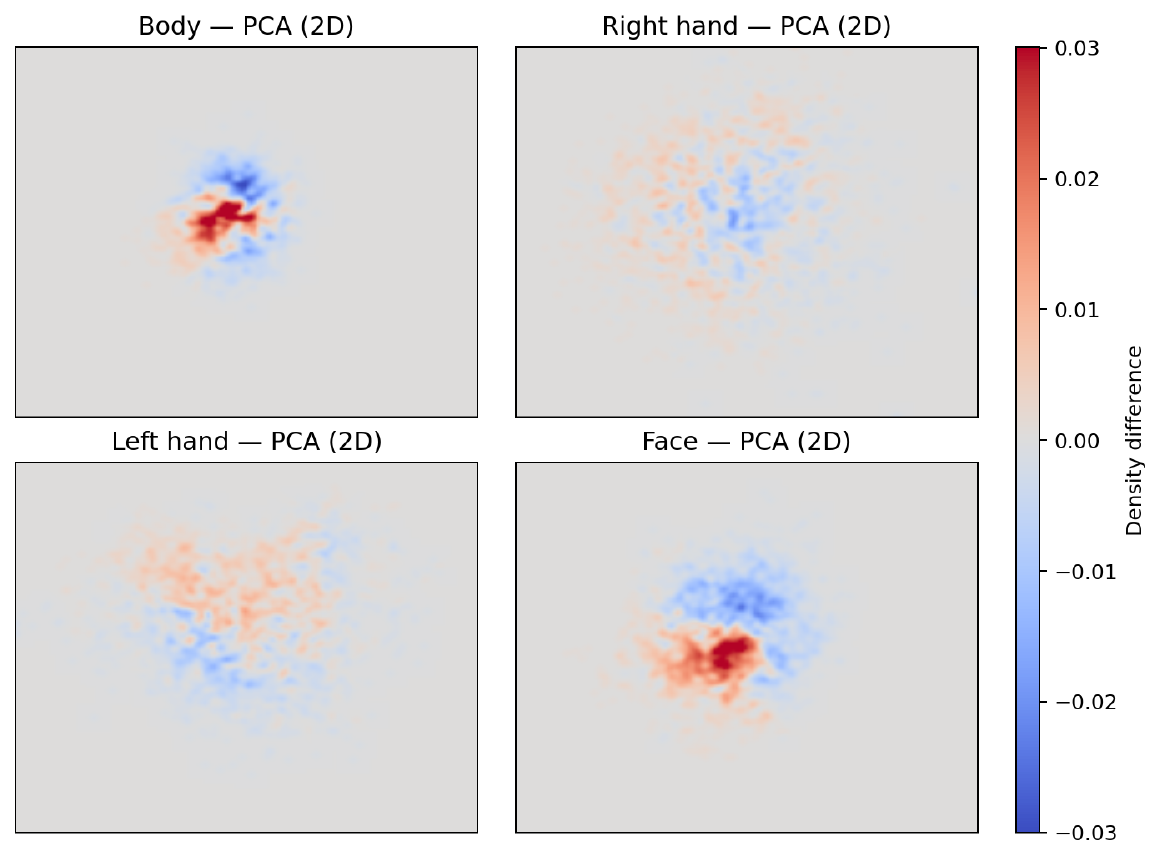}
        \vspace{1mm}
        \small\textbf{(d)} GT - Transformer w/o KL
    \end{minipage}
    \hfill
    \begin{minipage}[b]{0.46\columnwidth}
        \centering
        \includegraphics[width=\textwidth]{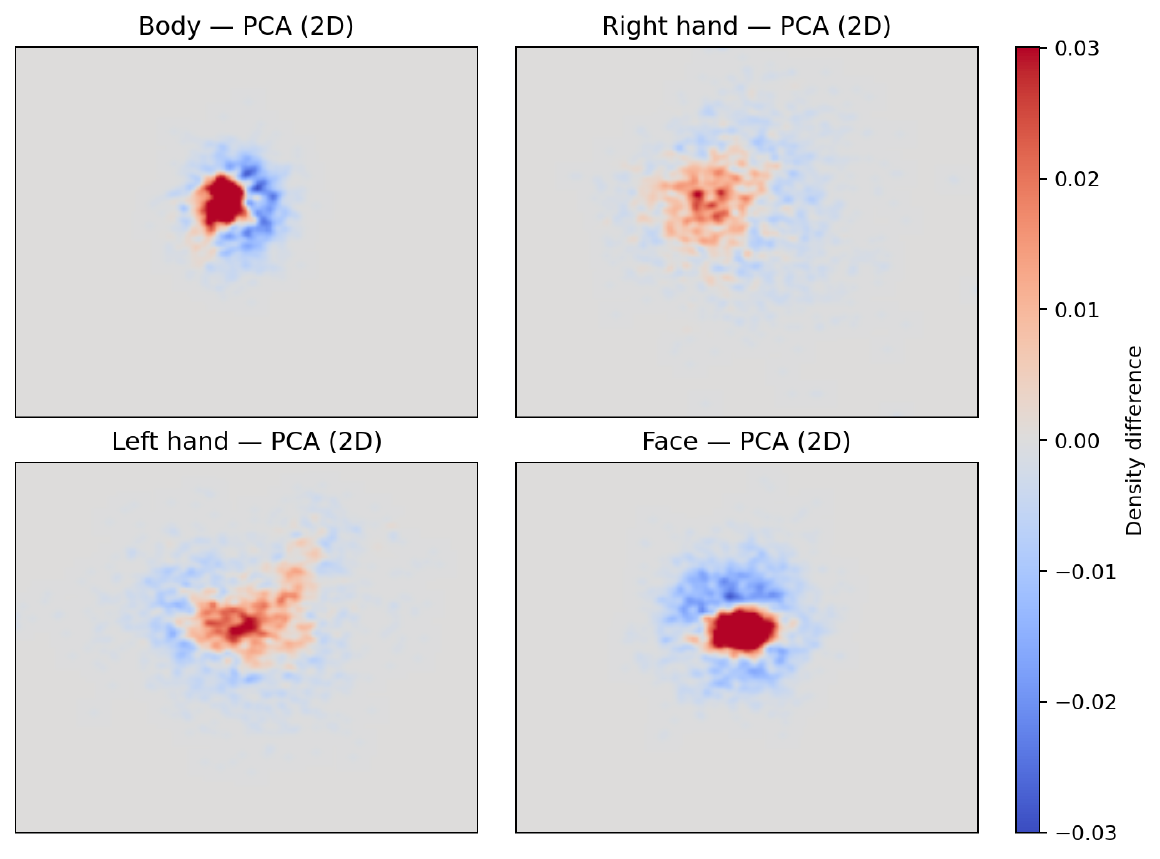}
        \vspace{1mm}
        \small\textbf{(e)} GT - Transformer w/ KL
    \end{minipage}

    \caption{(a–c) Regional PCA projections. 
    (d-e) Density-difference maps (GT – Transformer): red indicates areas with higher density than the AE, blue indicates lower. 
    }
    \label{fig:PCA_combined}
\end{figure}

As shown in Fig. \ref{fig:PCA_combined}(a), the AE produces compact, stable distributions while transformer without KL regularization produces more scattered and noisy patterns that deviate from the AE (Fig. \ref{fig:PCA_combined}(b)). Introducing KL regularization guides the Transformer's distributions toward the AE prior, reducing noise and enhancing compactness, particularly in the face and hand regions \ref{fig:PCA_combined}(c), indicating that KL improves the structural organization of the latent space.

To quantify the mismatch between the AE and the Transformer, we compute smooth 2D density estimates within each PCA plane and visualize the difference (GT minus Transformer) as heatmaps. Fig.~\ref{fig:PCA_combined}(c–d) illustrates the discrepancies between the two projected distributions. Without KL regularization, deviations are diffuse and scattered; particularly in the face and hand regions. When KL is applied, these differences become more centered and structured, suggesting that KL promotes a more coherent alignment of the Transformer's latent space with the AE prior.

\subsubsection{Statistical Analysis of the  Latent Spaces}

\begin{figure}[htbp]
    \centering

    \begin{minipage}[b]{0.49\textwidth}
        \centering
        \includegraphics[width=\textwidth]{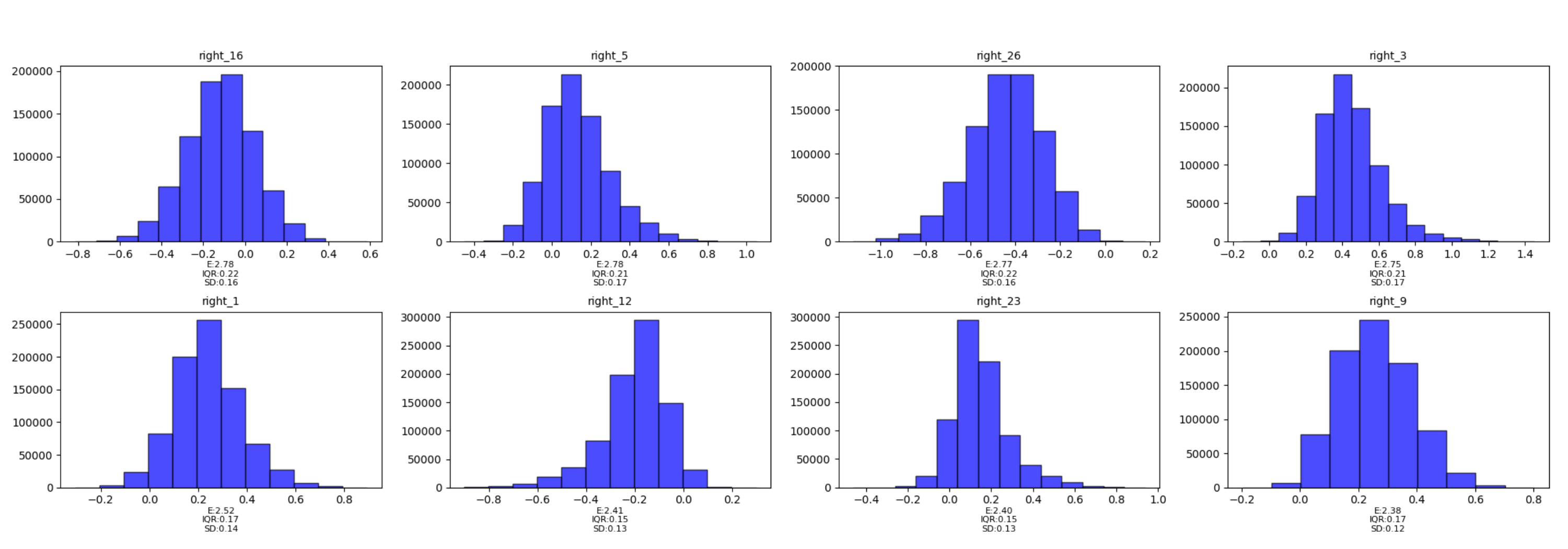}
        \vspace{1mm}
        \small\textbf{(a)} Right hand channels (avg. entropy: 2.01).
    \end{minipage}
    \hfill
    \begin{minipage}[b]{0.49\textwidth}
        \centering
        \includegraphics[width=\textwidth]{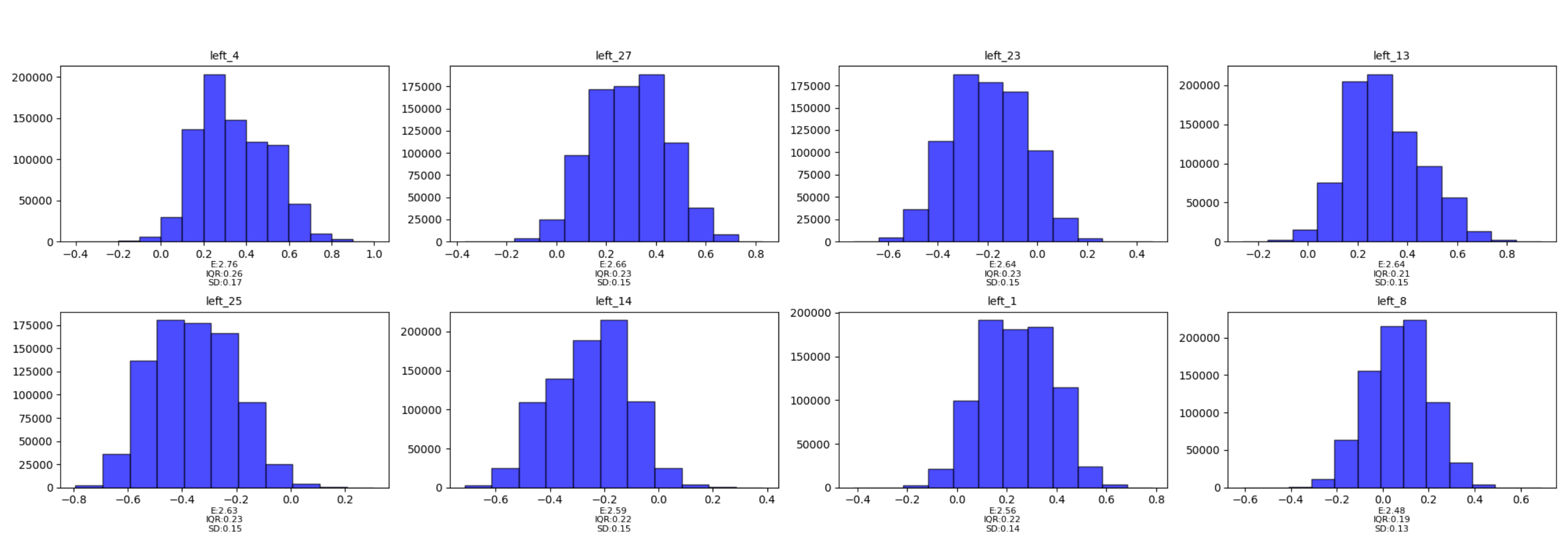}
        \vspace{1mm}
        \small\textbf{(b)} Left hand channels (avg. entropy: 2.22).
    \end{minipage}

    \vskip\baselineskip

    \begin{minipage}[b]{0.49\textwidth}
        \centering
        \includegraphics[width=\textwidth]{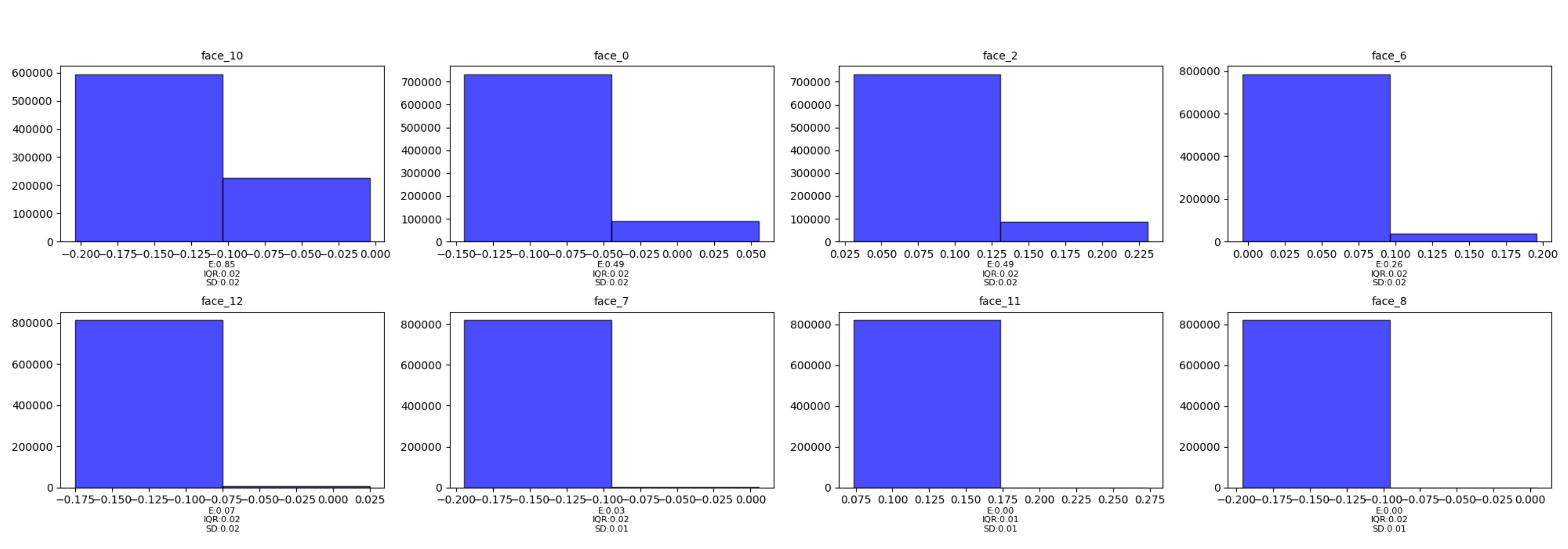}
        \vspace{1mm}
        \small\textbf{(c)} Face channels (avg. entropy: 0.14).
    \end{minipage}
    \hfill
    \begin{minipage}[b]{0.49\textwidth}
        \centering
        \includegraphics[width=\textwidth]{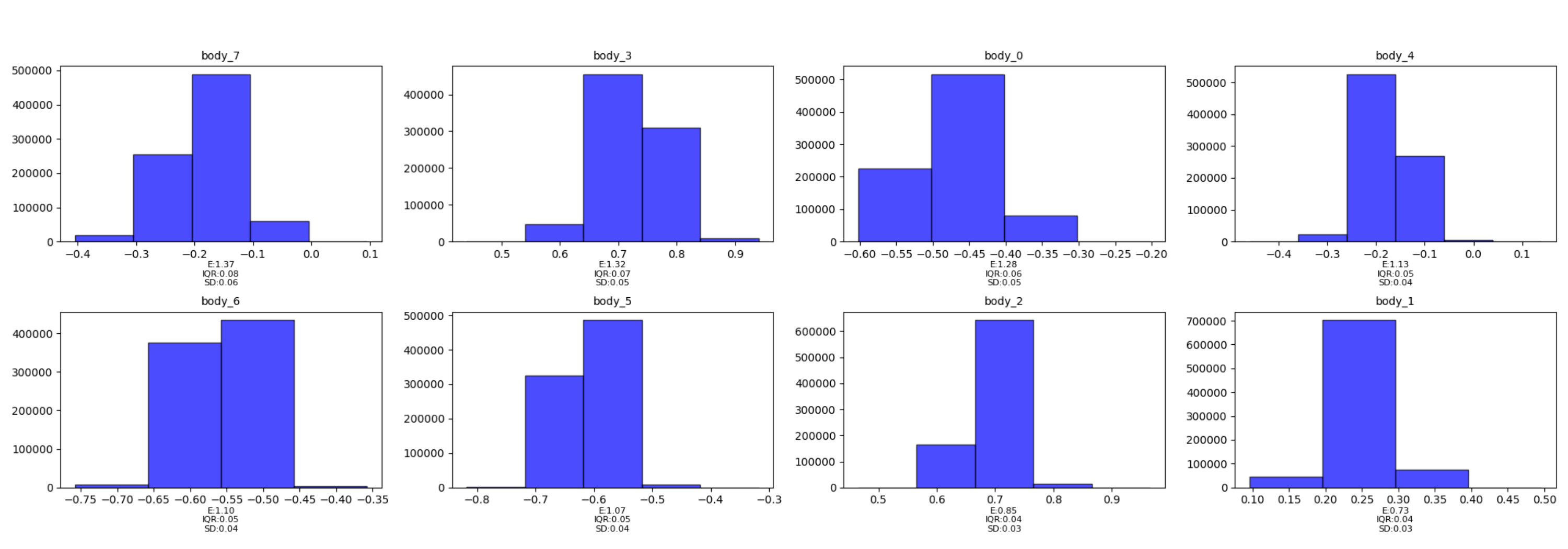}
        \vspace{1mm}
        \small\textbf{(d)} Body channels (avg. entropy: 1.11).
    \end{minipage}

    \caption{Histograms of top 8 latent channels with the highest entropy for each region of PHOENIX14T Dataset. Each subplot shows the distribution of a channel along with its entropy (E), interquartile range (IQR), and standard deviation (SD). (Zoom in for better visibility.)}
    \label{fig:Phoenix14t-channels}
\end{figure}

Further, statistical analysis of the learned latent representations reveals meaningful distinctions in the encoding behavior across regions for both dataset. In  Figure~\ref{fig:Phoenix14t-channels}, we present example histograms and corresponding entropy values for selected channel distributions from each structurally disentangled latent subspace (face, body, right hand, and left hand) learned by the autoencoder. 

\label{sec:learned-latent-space}

As can be seen, the face channels have minimal variance. In contrast, the right and left hand channels show broader, more dispersed distributions with higher entropy and standard deviation across most dimensions, confirming the rich variability and critical role of manual articulators in sign expression. The body encodings fall somewhere in between, reflecting more stable but still semantically relevant movement, however body region spans a larger physical space and thus generates higher magnitude latent activations, even when the underlying motion is less semantically dense. To mitigate this effect and maintain balanced learning, we increase the contribution of semantically richer regions by weighting. 

These distributional differences confirm that the latent space is semantically partitioned. They also highlight the need for targeted regularization and proper loss weighting. This encourages the model to focus more on dense, information-rich regions while still preserving reconstruction quality across all articulators.

\begin{figure}[htbp]
    \centering

    \begin{minipage}[b]{0.49\textwidth}
        \centering
        \includegraphics[width=\textwidth]{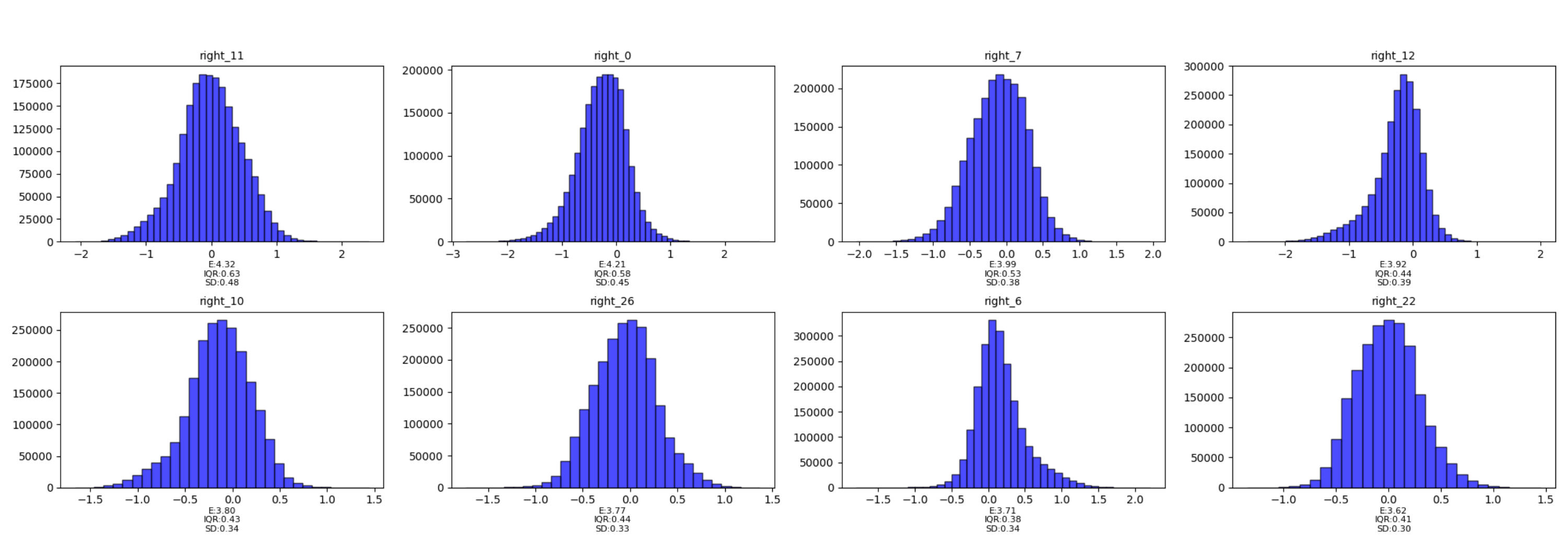}
        \vspace{1mm}
        \small\textbf{(a)} Right hand channels (avg. entropy: 2.98).
    \end{minipage}
    \hfill
    \begin{minipage}[b]{0.49\textwidth}
        \centering
        \includegraphics[width=\textwidth]{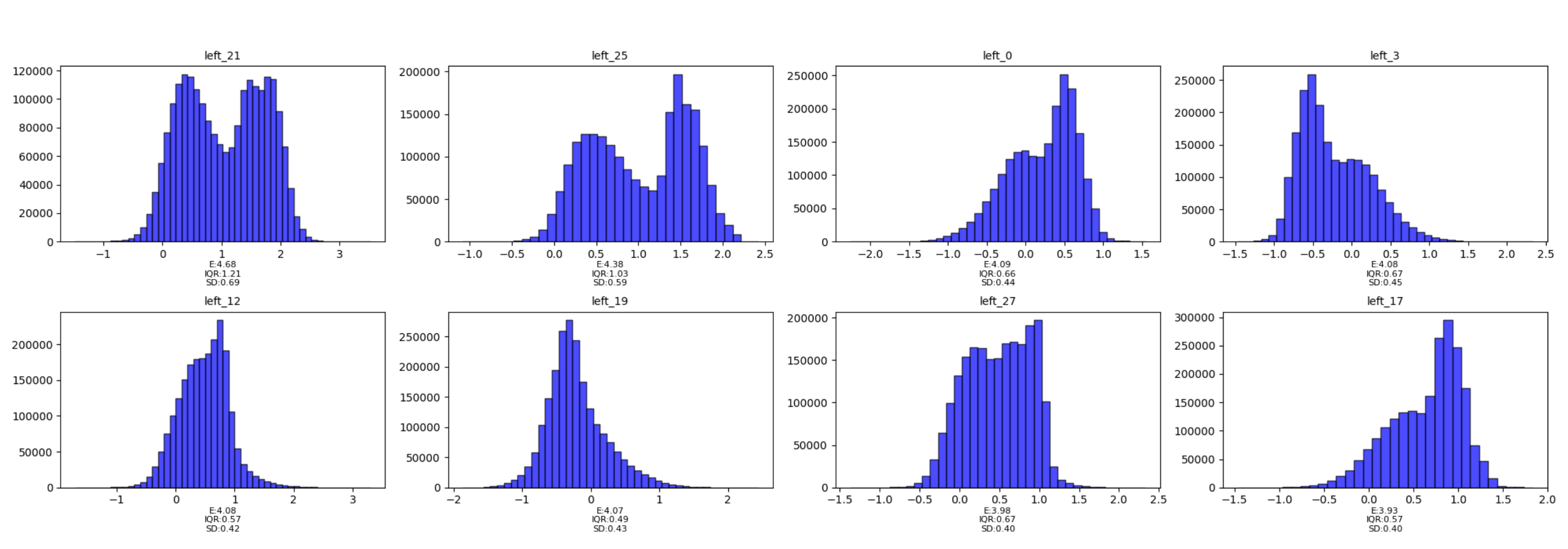}
        \vspace{1mm}
        \small\textbf{(b)} Left hand channels (avg. entropy: 3.19).
    \end{minipage}

    \vskip\baselineskip

    \begin{minipage}[b]{0.49\textwidth}
        \centering
        \includegraphics[width=\textwidth]{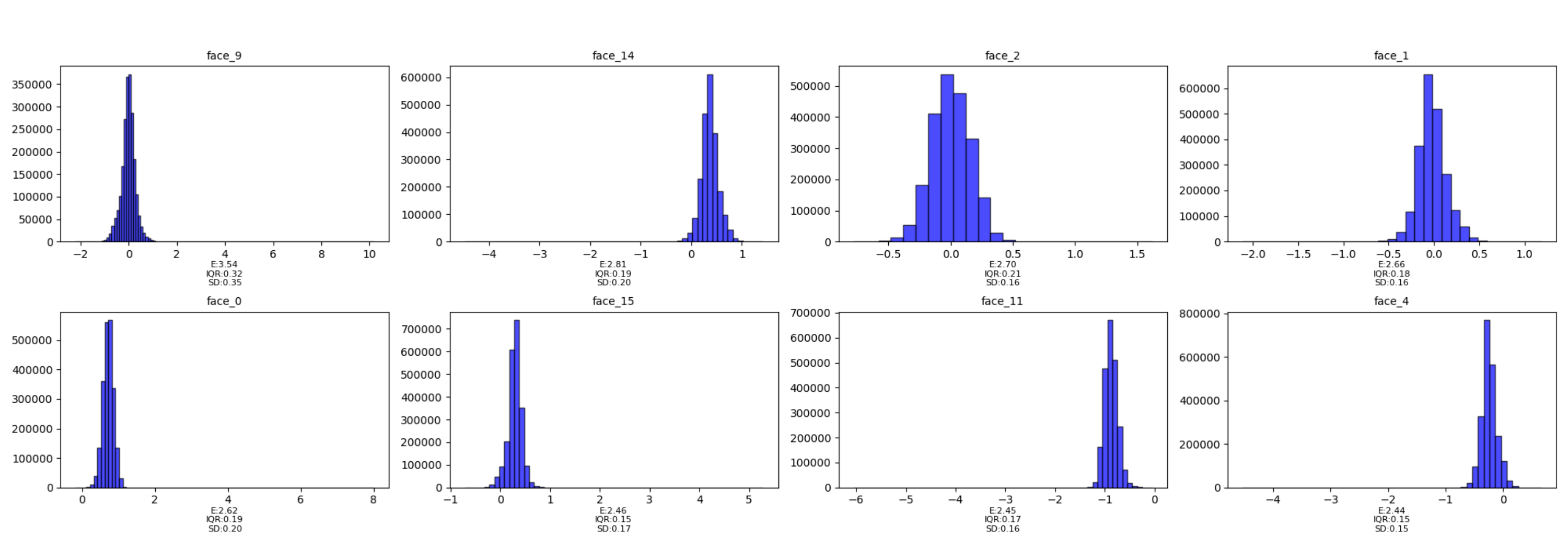}
        \vspace{1mm}
        \small\textbf{(c)} Face channels (avg. entropy: 2.21).
    \end{minipage}
    \hfill
    \begin{minipage}[b]{0.49\textwidth}
        \centering
        \includegraphics[width=\textwidth]{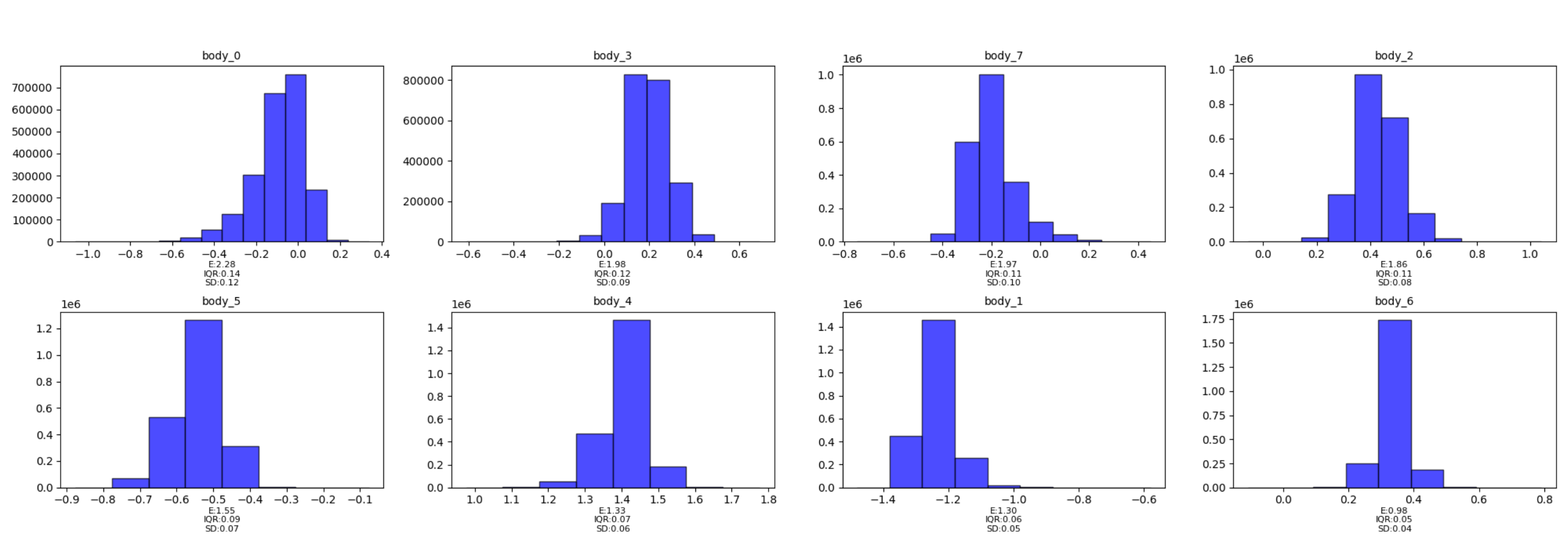}
        \vspace{1mm}
        \small\textbf{(d)} Body channels (avg. entropy: 1.65).
    \end{minipage}

    \caption{Histograms of top 8 latent channels with the highest entropy for each region of CSL-Daily Dataset. Each subplot shows the distribution of a channel along with its entropy (E), interquartile range (IQR), and standard deviation (SD). (Zoom in for better visibility.)}
    \label{fig:CSL-channels}
\end{figure}

The latent space structure observed in CSL-Daily mirrors that of PHOENIX14T, with clear separation between articulator groups and similar relative entropy patterns, manual articulators exhibit the highest variability, followed by face and body (Figure \ref{fig:CSL-channels}). CSL-Daily encodings exhibit broader distributions across all articulator groups, as reflected by higher entropy values, particularly in the manual articulators. The left and right hand channels display the widest spread, with average entropies around 3.0, indicating increased variability and expressiveness in CSL’s daily conversational context. Even face channels, which previously showed limited variance, now present more active and information-rich distributions. This broader activation aligns with CSL-Daily’s more diverse signer pool and open-domain language, reinforcing the need for careful latent space design.

\subsection{Test Results}

Table~\ref{tab:results} reports test set results on PHOENIX14T, comparing our method with recent autoregressive and non-autoregressive approaches. 
The DARSLP baseline already outperforms most prior work without requiring gloss annotations, demonstrating the benefit of disentangled latent encodings. 
Adding KL divergence during the second training phase further improves generalization, boosting BLEU-4 from 8.72 to 11.07 and ROUGE from 23.25 to 32.55. 
This confirms that aligning predicted embeddings with learned priors reduces overfitting to dominant high-variance features (e.g., body motion) and yields smoother, more accurate sign pose sequences.

\begin{table}[h]
\centering
\scriptsize
\setlength{\tabcolsep}{2.8pt} 
\begin{tabular}{l|c|c c c c c c c}
\toprule
\textbf{Model} & \textbf{G} & \textbf{DTW↓} & \textbf{B1↑} & \textbf{B2↑} & \textbf{B3↑} & \textbf{B4↑} & \textbf{chrF↑}  & \textbf{ROUGE↑}\\
\midrule
GT \textsuperscript{\dag} & - & 0.000 & 34.43 & 22.08 & 16.13 & 12.81 & 34.62 & 35.22\\
\midrule
\multicolumn{9}{c}{\textit{Autoregressive}} \\ 
\midrule
PT Base \cite{walsh2024signstitchingnovelapproach} & \cmark & 0.197 & 6.27 & 3.33 & 2.14 & 1.59  & 9.50 & 14.52 \\ 
PT FP\&GN \cite{walsh2024signstitchingnovelapproach} & \cmark & 0.191 & 11.45 & 7.08 & 5.08 & 4.04 & 19.09 & 14.52 \\
G2P-MP \cite{xie2023g2pddmgeneratingsignpose} & \cmark & 0.146 & 15.43 & 10.69 & 8.26 & 6.98 & - & - \\
G2P-DDM \cite{xie2023g2pddmgeneratingsignpose} & \cmark & 0.116 & 16.11 & 11.37 & 9.22 & 7.50 & - & - \\
SignVQNet \cite{hwang2024autoregressivesignlanguageproduction} & \xmark & 0.299 & - & - & - & 6.88 & - & - \\
Stitching T2P \cite{walsh2024signstitchingnovelapproach} & \xmark & 0.572 & 25.14 & 13.54 & 8.98 & 6.67 & 29.50 & 26.49 \\
GCDM \cite{10.1145/3663572} & \cmark & - & 22.03 & 14.21 & 10.16 & 7.91 & 23.20 & - \\
\midrule
\multicolumn{9}{c}{\textit{Non-Autoregressive}} \\
\midrule
NSLP-G \cite{hwang2024autoregressivesignlanguageproduction} & \xmark & 0.638 & - & - & - & 5.56 & - & - \\
NSLP-G w/o FT \cite{hwang2024autoregressivesignlanguageproduction} & \xmark & 0.646 & - & - & - & 4.41 & - & - \\
NAT-AT \cite{10.1145/3474085.3475463} & \cmark & 0.177 & 14.26 & 9.93 & 7.11 & 5.53 & 21.87 & 18.72 \\
NAT-EA \cite{10.1145/3474085.3475463} & \cmark & 0.146 & 15.12 & 10.45 & 7.99 & 6.66 & 22.98 & 19.43 \\
NSVQ + Non-AR \cite{walsh2024datadrivenrepresentationsignlanguage} & \xmark & 0.105 & 27.74 & 16.36 & 11.75 & 9.20 & - & 27.93 \\
\midrule
\textbf{DARSLP (Ours)} & \xmark & 0.0439 & 24.97 & 15.91 & 11.35 & 8.72 & 29.67 & 23.25 \\
\textbf{DARSLP + KL (Ours)} & \xmark & \textbf{0.0394} & \textbf{33.17} & \textbf{20.47} & \textbf{14.38} & \textbf{11.07} & \textbf{31.69} & \textbf{32.55} \\
\bottomrule
\end{tabular}
\caption{PHOENIX14T TEST set comparison. Scores via back translation model \cite{walsh2025slrtp}.}
\label{tab:results}
\end{table}

\begin{figure}[h]
    \centering

    \begin{minipage}[c]{0.04\textwidth}
        \adjustbox{valign=m}{\rotatebox{90}{\small a) Predicted}}
        \vspace{0.35cm}

        \adjustbox{valign=m}{\rotatebox{90}{\small  b) Ground Truth}}
    \end{minipage}%
    \begin{minipage}[c]{0.44\textwidth}
        \includegraphics[width=\textwidth]{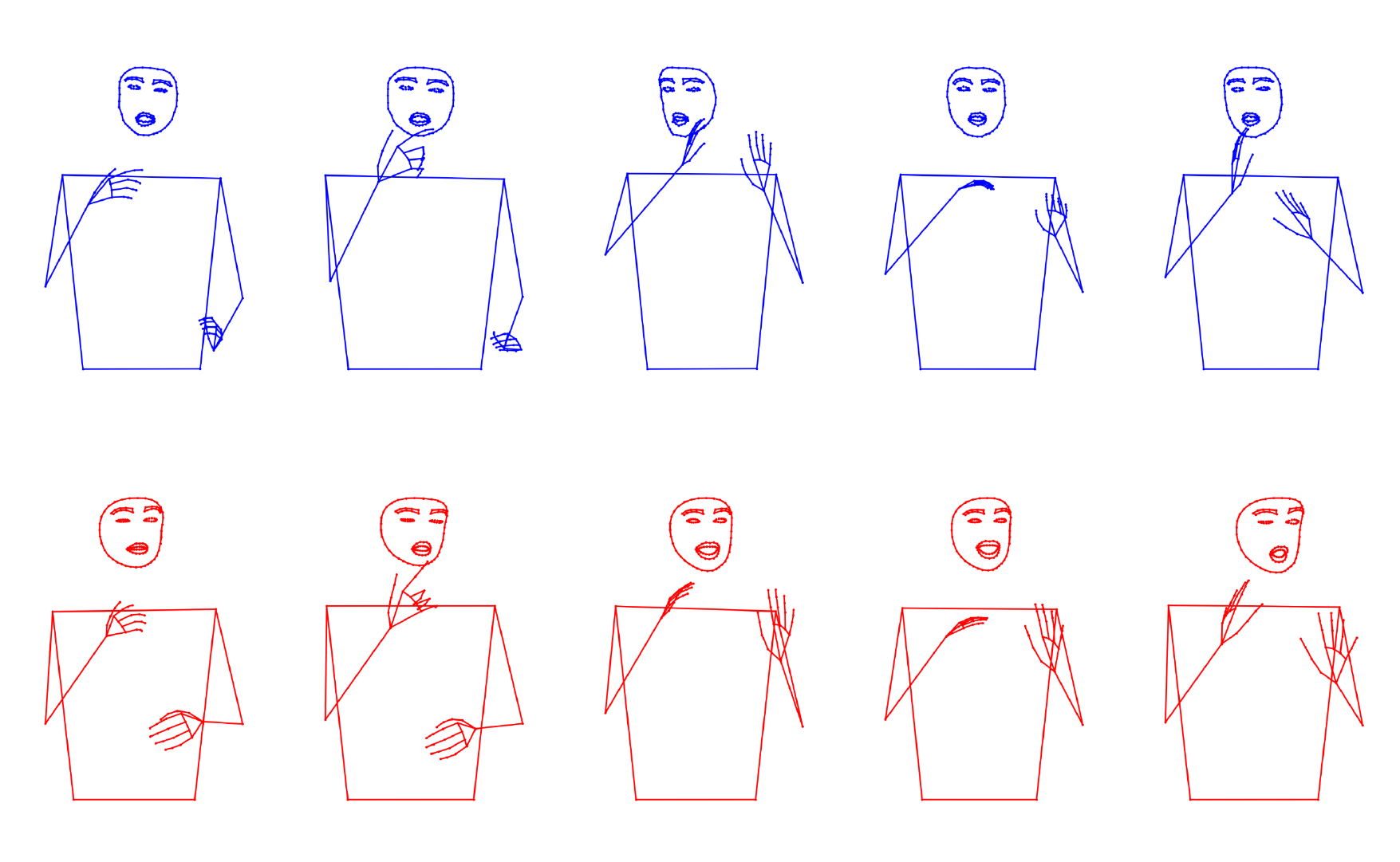}
    \end{minipage}

    \vspace{0.1cm}

    \begin{minipage}[c]{0.04\textwidth}
        \adjustbox{valign=m}{\rotatebox{90}{\small a) Predicted}}
        \vspace{0.35cm}

        \adjustbox{valign=m}{\rotatebox{90}{\small b) Ground Truth}}
    \end{minipage}%
    \begin{minipage}[c]{0.44\textwidth}
        \includegraphics[width=\textwidth]{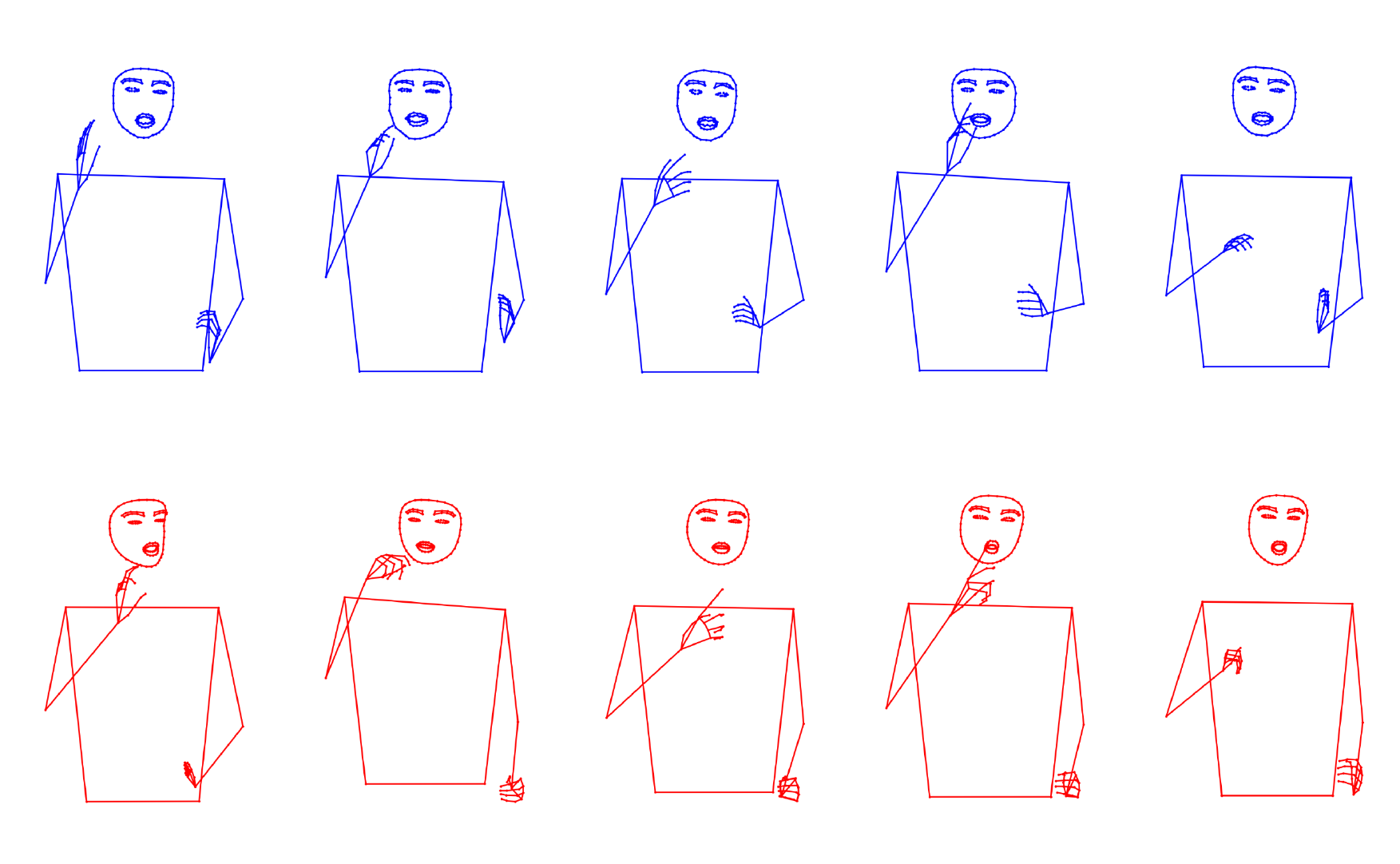}
    \end{minipage}

    \caption{\textbf{Pose sequences generated from input sentences.}  
    Top: \textit{"am mittwoch zieht von der nordsee regen heran der sich am donnerstag ausbreitet"}.  
    Bottom: \textit{"und nun die wettervorhersage für morgen freitag den vierundzwanzigsten juli"}.}
    \label{fig:pose_comparison}
\end{figure}

Qualitative examples in Figure~\ref{fig:pose_comparison} show that the model captures key manual and non-manual articulations. While the outputs are not frame-perfect reproductions, they generally preserve the semantic intent and flow of the original signing. Despite relying on a non-autoregressive decoder, the model is able to produce smooth transitions and plausible motion patterns, suggesting that the proposed representation and training approach effectively support expressive sign generation. These results, achieved without gloss supervision or large pretrained models, demonstrate the effectiveness and efficiency of our proposed framework under limited supervision.

\textbf{Generalization to CSL-Daily.} We evaluate our finalized transformer model, selected on PHOENIX14T, on the CSL-Daily dataset using the same architecture and training strategy. To accommodate CSL-Daily's broader linguistic context and higher variance, we added another projection layer when mapping keypoints to the 80-dimensional latent space. This produced latent distributions, particularly for the right hand, that aligned more closely with the assumed Gaussian prior, making KL regularization more effective (see Supplementary Material for details). As shown in Table~\ref{tab:csl-results}, the model with KL divergence regularization outperforms the variant trained without KL across all metrics. Despite the domain shift, our model achieves competitive BLEU and ROUGE scores relative to ground-truth, which sets an upper-bound, demonstrating strong generalization to new language content and signer variability. As no prior SLP results exist for sign pose generation task on this dataset, we compare our performance against the ground truth (GT) baseline.

\begin{table}[h]
    \centering
    \footnotesize
    \renewcommand{\arraystretch}{1.1}
    \setlength{\tabcolsep}{2.5pt} 
    \begin{tabular}{l|c c c c c c c}
        \toprule
        \textbf{Model} & \textbf{DTW-MJE↓} & \textbf{B1↑} & \textbf{B2↑} & \textbf{B3↑} & \textbf{B4↑} & \textbf{chrF↑} & \textbf{R↑}  \\
        \midrule
        GT & 0.000 & 21.81 & 10.90 & 6.31 & 4.02 & 5.26 & 21.63 \\
        \midrule
        \textit{DARSLP} & 0.177 & 18.72 & 8.33 & 4.46 & 2.73 & 4.24  & 19.13\\
        \textit{DARSLP+KL} & \textbf{0.175} & \textbf{20.47} & \textbf{9.55} & \textbf{5.45} & \textbf{3.53} & \textbf{4.89} & \textbf{20.55} \\
        \bottomrule
    \end{tabular}
    \caption{Evaluation results on the CSL-Daily TEST set.}
    \label{tab:csl-results}
\end{table}

\section{Conclusions and Future Work}

We proposed a gloss-free sign language production framework that combines context-aware text encoding, a semantically structured latent pose space, and dynamic temporal control within a unified architecture. 
Despite not relying on gloss annotations or large pretrained generative models, the proposed approach achieves competitive performance through a compact design and targeted training strategy. 

Beyond its practical performance, this framework offers a promising research direction by enabling interpretable analysis of region-wise latent behavior. Our visualizations demonstrate how the model responds to variance in specific articulators, offering key insights into the internal dynamics of sign language generation. To the best of our knowledge, this is the first work to present such a targeted, interpretable latent space analysis in the context of sign language production. As such, it also contributes to the broader goal of interpretable AI in sequence generation models.

While qualitative results demonstrate smooth and plausible motions, there is still room to improve visual fidelity and expressiveness. Current evaluation of SLP models relies heavily on back-translation, which introduces additional uncertainty. More direct and reliable evaluation techniques are needed to assess generation quality more faithfully. In parallel with improving evaluation, further architectural refinements and training strategies are also needed. Although this study adopts a region-wise latent structure and disentangled modeling across articulators, assigning appropriate loss weights remains challenging, as each region exhibits different levels of motion complexity and semantic importance. Future work will explore adaptive weighting schemes that better reflect joint-level contributions, particularly to capture the hierarchical motion structure of hands more effectively.

\clearpage